\pdfoutput=1
\documentclass[runningheads]{llncs}
\usepackage{graphicx}
\usepackage{amsmath,amssymb} % define this before the line numbering.
\usepackage{color}
\usepackage[ruled,vlined]{algorithm2e}
\usepackage{multirow}
\usepackage{mathrsfs}
\usepackage{epsfig}
\usepackage{epstopdf}
\usepackage[width=122mm,left=12mm,paperwidth=146mm,height=193mm,top=12mm,paperheight=217mm]{geometry}
\begin{document}

\pagestyle{headings}
\mainmatter

\title{Exploring Linear Relationship in Feature Map Subspace for ConvNets Compression} % Replace with your title

\titlerunning{ConvNets Compression}

\authorrunning{Wang \emph{et al.}}

\author{Dong Wang$^1$, Lei Zhou$^1$, Xueni Zhang$^1$, Xiao Bai$^1$, and Jun Zhou$^2$}

\institute{$^1$Beihang University $^2$Griffith University}

\maketitle

\begin{abstract}
While the research on convolutional neural networks (CNNs) is progressing quickly, the real-world deployment of these models is often limited by computing resources and memory constraints. In this paper, we address this issue by proposing a novel filter pruning method to compress and accelerate CNNs. Our work is based on the linear relationship identified in different feature map subspaces via visualization of feature maps. Such linear relationship implies that the information in CNNs is redundant. Our method eliminates the redundancy in convolutional filters by applying subspace clustering to feature maps. In this way, most of the representative information in the network can be retained in each cluster. Therefore, our method provides an effective solution to filter pruning for which most existing methods directly remove filters based on simple heuristics. The proposed method is independent of the network structure, thus it can be adopted by any off-the-shelf deep learning libraries. Experiments on different networks and tasks show that our method outperforms existing techniques before fine-tuning, and achieves the state-of-the-art results after fine-tuning.

\end{abstract}

\section{Introduction}

With the collection of huge volume of labeled data, tremendous power of graphical processing units (GPUs) and parallel computation, convolutional neural networks (CNNs) have achieved the state-of-the-art performance in a wide variety of computer vision tasks, such as image classification~\cite{he2016deep}, object detection~\cite{ren2015faster}, image segmentation~\cite{he2017mask}, and human pose estimation~\cite{newell2017associative}. As flexible function approximators by scaling to millions of parameters, CNNs can extract high-level and more discriminative features compared with the traditional elaborative hand-crafted ones.

However, modern CNNs heavily rely on the intensive computing and memory resources despite their overwhelming success. For instance, the ResNet-50~\cite{he2016deep} has more than 50 convolutional layers, requiring over 95MB storage memory and over 3.8 billion floating number multiplications when processing an image. The VGG-16 model~\cite{simonyan2014very} has 138.34 million parameters, taking up more than 500MB storage space, and needs 30.94 billion float point operations (FLOPs) to classify a single image. It is very difficult to deploy these complex CNN models in some specific scenarios where computing resource is constrained, i.e., a task must be completed with limited resources such as computing time, storage space, and battery power.

Both academia and industry have developed methods to reduce the amount of parameters in CNNs. Ba \emph{et al.}~\cite{ba2014deep} used class probabilities produced by a pre-trained model as ``soft targets" to feed a tiny network, successfully transferring the cumbersome model to a compact one while maintaining the generalization capability of the model. The student-teacher paradigm in~\cite{ba2014deep} has shown its effectiveness in compressing CNNs, however, to devise a new tiny network is not a trivial task. Moreover, it remains an open problem on how to define the inherent ``knowledge" in the teacher model. Tensor factorization based methods~\cite{lebedev2014speeding,kim2015compression,jaderberg2014speeding,gong2014compressing} factorize an over-parameterized convolutional layer into several light ones. However, decomposing $1\times 1$ convolution favoured by modern CNNs (e.g., GoogleNet~\cite{szegedy2015going}, ResNet~\cite{he2016deep}, and Xception~\cite{chollet2016xception}) is still an intractable problem. Moreover, tensor decomposition techniques expand the target network deeper, incurring more convolution operations.

Filter pruning has been proposed to address this issue. Since the network architecture is constant after the filter pruning, the obtained model is compatible with any off-the-shelf deep learning frameworks. In addition, since volumes of both convolutional kernels and intermediate activations are shrunken, the required memory is reduced remarkably. This strategy also allows complementary compression methods to be employed to gain a more compact model. The advantages of filter pruning lead to increasing attention to research in this direction. He \emph{et al.}~\cite{he2017channel} learned a sparse weight vector to measure the importance of filters by applying the LASSO regression. Luo \emph{et al.}~\cite{luo2017thinet} used statistical information computed from the next layer to guide the filter pruning for the current layer in a greedy way. Both methods directly prune some filters. It is obvious that the information contained in pruned filters can no longer be utilized once the filters have been pruned.

\begin{figure}[htb!]
\centering
\includegraphics[width=1\linewidth]{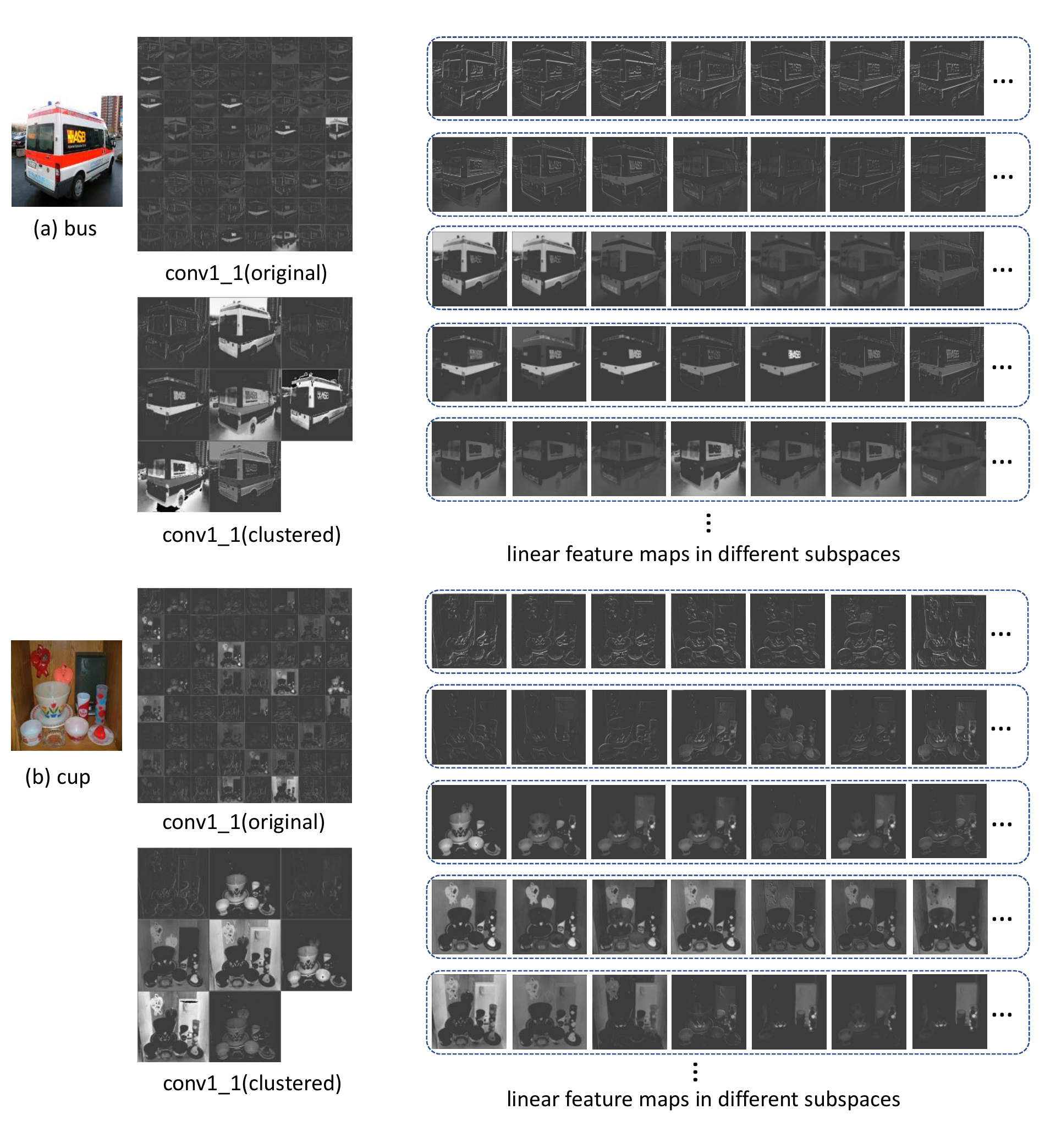}
\caption{Visualization of output feature maps produced by the first convolutional layer of VGG-16~\cite{simonyan2014very}. The example images are randomly chosen from ILSCVR-12~\cite{russakovsky2015imagenet}. We show the linear feature maps in different subspaces and the feature maps produced by 8 clustered convolutional filters. It is obvious that redundant feature maps can be eliminated through subspace clustering algorithm.}
\label{featuremap}
\end{figure}

In this paper, we propose a novel filter pruning method by leveraging the linear relationship in feature map subspace. As shown in Figure~\ref{featuremap}, since different feature maps from one convolutional layer originate from the same image with different convolutional filters by a linear operation, the outputs will be linearly dependent in different subspaces. Among various subspace analysis approaches, subspace clustering is an excellent method to cluster linearly distributed data in different subspaces. Motivated by this, instead of measuring the importance of filters~\cite{li2016pruning,hu2016network} or feature maps~\cite{he2017channel,luo2017thinet,anwar2016compact} and subsequently removing the trivial kernels, we attempt to seek the most representative information by casting the filter selection problem into a subspace clustering problem on intermediate activations. We firstly cluster feature maps into subspaces. This allows the clustering of the corresponding filters in the next layer which take these feature maps as input. Also, filters in the upper layer that produce these feature maps can be clustered. Then, we iterate this process to prune the whole network layer by layer.

In summary, this paper makes the following contributions:
\begin{itemize}	
\item We propose a novel filter pruning method based on the linear relationship in feature map subspace to compress and accelerate CNN models. To the best of our knowledge, it is the first work to investigate clustering method for CNN model compression. Moreover, it is also the first work to employ subspace clustering to accelerate CNNs. We can prune the redundant information in feature maps and simultaneously retain the most representative information.
\item We devise a flexible filter pruning framework that is independent of the network structure. Thus our method can be well supported by any off-the-shelf deep learning libraries.
\item Compared to the original heavy network, experiments demonstrate that the learned portable network achieves a comparable accuracy, but has significantly lower memory usage and computational cost. We achieve consistent improvement on various tasks, exceed other filter pruning works, and obtain the state-of-the-art results.
\end{itemize}

\section{Related Work}

Recently, several works on CNN acceleration focused on network pruning thanks to its apparent benefits. Along this line, various strategies have been adopted, e.g., fine-grained pruning~\cite{han2015learning}, group-level pruning~\cite{wen2016learning,lebedev2016fast}, filter-level pruning~\cite{hu2016network,li2016pruning}, layer-level pruning~\cite{zagoruyko2016paying} and feature maps pruning~\cite{figurnov2016perforatedcnns,anwar2016compact,he2017channel,luo2017thinet}. Han \emph{et al.}~\cite{han2015learning} introduced sparsity regularization approach to calculate and remove connections with small weights. The major drawback of this fine-grained pruning is the loss of universality and flexibility due to the unstructured pruned parameters, which heavily hinders the pruned models to be transferred to real applications.

Group-level pruning approaches alleviate the above problem by learning solid sparse patterns. Lebedev \emph{et al.}~\cite{lebedev2016fast} used group-wise brain damage process to sparsify convolution kernels. This generates one sparsity pattern per group (2D kernels) in convolutional layers. Then the entire group with small weights can be removed. Similarly, Wen \emph{et al.}~\cite{wen2016learning} proposed the Structured Sparsity Learning (SSL) method to regularize filter, channel, filter shape and depth structures. Zagoruyko \emph{et al.}~\cite{zagoruyko2016paying} demonstrated a layer-level pruning technique. For the network consisting of multiple homogeneous stages (each stage is a set of convolutional blocks), some stages are removed by merging attention maps into specific cost function. Thus, it provides an approach to combine network pruning with knowledge distillation. Figurnov \emph{et al.}~\cite{figurnov2016perforatedcnns} explored feature map pruning which only kept a subset of rows in the patch matrix by using solid sparsity patterns, i.e., perforation mask~\cite{figurnov2016perforatedcnns}, and interpolated the missing output values. Perforation mask was predefined and could be in grid or pooling structure. However, this method only shortens the inference time and does not compress the model. Similar to~\cite{lebedev2016fast}, it is only supported by deep learning frameworks which reduce generalized convolution to matrix multiplication.

Compared with the aforementioned pruning strategies, filter-level pruning is more efficient in accelerating very deep neural networks. For two consecutive convolutional blocks, which are indispensable in all CNNs, after the filter pruning for the former block, the number of input channels of the latter block is also reduced. Moreover, the shape of the chunk created by the latter block is constant. By minimizing construction errors of feature maps, the outputs of CNN endpoints are retained. Thus, it is vital to determine which filters are to be eliminated. Some methodologies use kernel importance. An intuitive possible way is to use the magnitude of weights. Li \emph{et al.}~\cite{li2016pruning} calculated absolute weight sum of each filter as its importance score. Denseness in filters is an alternative. Hu \emph{et al.}~\cite{hu2016network} depicted the significance of each filter by calculating the Average Percentage of Zeros (APoZ) in it.

Another type of approaches conquers the filter selection challenge by converting it into channel selection for feature maps. Anwar \emph{et al.}~\cite{anwar2016compact} exhibited over 100 random trials on channel selection. However, it is time consuming and laborious per trial. Thus, inflexibility is a common problem on very deep models and large datasets. ThiNet~\cite{luo2017thinet} used statistical information computed from the next layer to guide the filter pruning of the current layer. The pruned convolutional layer was forced to mimic the original one by minimizing the reconstruction errors of feature maps. Although the works from He \emph{et al.}~\cite{he2017channel} and Luo \emph{et al.}~\cite{luo2017thinet} have similar workflows, their channel selection strategies are different. The main insight of~\cite{he2017channel} was to learn a weight vector for feature maps. The weight vector is optimized for channel selection with fixed convolutional filters. Then the convolutional filters are used to reconstruct error with the weight vector fixed. In practice, the weight vector is optimized for multiple times and the filters just once to obtain the final result since the two step iteration is time consuming.

In addition, there are a variety of techniques for compressing convolution filters. A representative approach is Low-rank approximation~\cite{lebedev2014speeding,kim2015compression,jaderberg2014speeding,gong2014compressing}. It breaks a convolutional layer into several small pieces by applying tensor decomposition strategies, e.g. CP decomposition~\cite{lebedev2014speeding} and Tucker decomposition~\cite{kim2015compression}. Other methods include parameter quantization~\cite{gong2014compressing,chen2015compressing,wu2016quantized,han2015deep} and structural matrix design~\cite{wang2017beyond,cheng2015exploration,sindhwani2015structured}.

\section{Proposed Method}

In this section, we describe a novel filter pruning method based on the linear relationship in feature map subspace. We first introduce the overall framework, then present the details of each step. Finally, we describe our pruning strategy which takes both efficiency and effectiveness into consideration.

\subsection{Overall Framework}

\begin{figure}[t]
\centering
\includegraphics[width=0.9\linewidth]{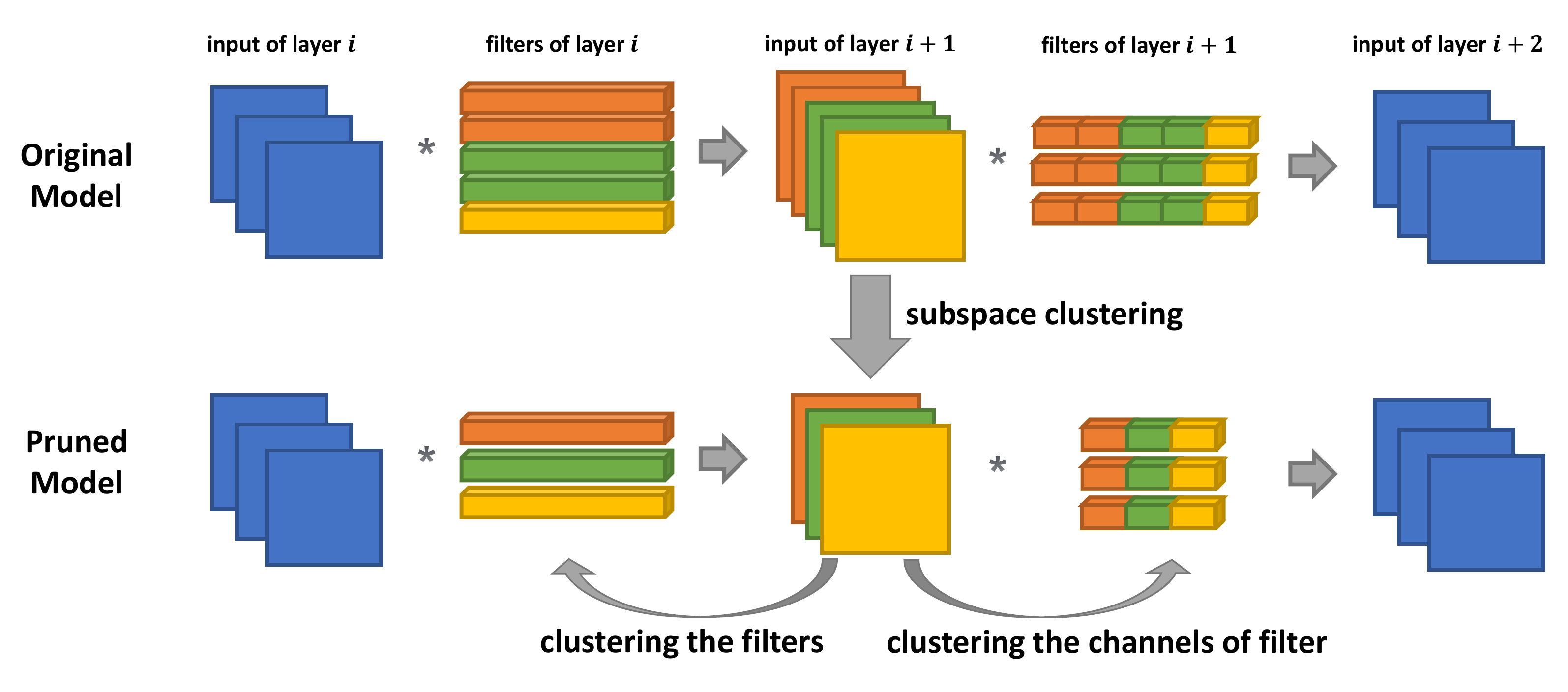}
\caption{Illustration of our filters pruning method. First, we cluster the input feature maps of layer $i+1$ by a subspace clustering algorithm. Then the filters in layer $i$ and the channels of filters in layer $i+1$ can be pruned by respectively calculating the average of corresponding filters and channels of each feature map cluster.}
\label{pruning}
\end{figure}

Filter pruning is an effective method for reducing the complexity of neural networks. There are two key points in filter pruning. The first is filter selection, i.e., we need to seek the most representative filters to retain as much information as possible. The second is reconstruction, i.e., the following feature maps shall be reconstructed using the clustered filters. The main difference between our method and previous works is the strategy in filter selection. Most existing methods directly prune filters which make weak contribution to the neural network. The drawback is that the information of pruned filters can not be further utilized, which influences the result of feature map reconstruction. Our method, on the other hand, utilizes subspace clustering algorithm to reduce the number of feature maps, which can simultaneously eliminate the redundant feature maps and retain the most representative information in feature maps. To prune the input feature maps from $c$ to desired $c^{\prime}$ $(0\le c^{\prime} \le c)$, we group the $c$ feature maps into $c^{\prime}$ clusters. Then we calculate the average of corresponding filters of each feature map cluster. A pre-trained model is pruned layer by layer with a predefined compression rate.

We summarize our filter pruning method on a single convolutional layer in Figure~\ref{pruning}. We aim to prune the filters in layer $i$ and layer $i+1$. Once feature maps of layer $i+1$ are clustered, we can cluster corresponding filters in layer $i$ and corresponding channels of filter in layer $i+1$. The method has the following key steps:
\begin{enumerate}
  \item \textbf{Feature map clustering.} Since different feature maps from a convolutional layer are generated from the same image with different convolutional filters by a linear operation, the output feature maps will be linearly dependent in different subspaces. Therefore, we leverage a subspace clustering algorithm to cluster the feature maps.
  \item \textbf{Filter clustering and reconstruction.} After subspace clustering, we can cluster corresponding input channels of filter in the next layer which take these feature maps as input. Filters in the upper layer that produce these feature maps can also be clustered. Then we reconstruct the following feature maps using the pruned filters. Note that the pruned network has exactly the same structure but with fewer filters. In other words, the original thick network becomes a much thinner model.
  \item \textbf{Fine-tuning.} Fine-tuning is a necessary step to recover the generalization ability influenced by filter pruning, which is time consuming on large datasets and complex models. For efficiency considerations, we fine-tune part of epochs after all pruned feature maps have been reconstructed.
  \item \textbf{Iterate to step 1 to prune the next layer.}
\end{enumerate}

\subsection{Filter Pruning}

we propose a two-step algorithm for filter pruning. In the first step, we aim to seek the most representative filters. Since there is a linear relationship in different feature map subspace, we utilize a subspace clustering algorithm to estimate average feature maps which contain as much representative information as possible. Then, the corresponding filters of each feature map cluster are clustered. In the second step, we reconstruct the following feature maps using the average filters with linear least squares.

Formally, we use $(X^{(i)}, W^{(i)}, *)$ to denote the convolution process in layer $i$, where $X^{(i)}\in \mathbb{R}^{c_i\times H\times W}$ is the input tensor which has $c_i$ feature maps of $H\times W$ in size. $W^{(i)}\in \mathbb{R}^{c_{i+1}\times c_i\times k_h\times k_w}$ is a set of filters with kernels of $k_h\times k_w$ in dimension, which generates a new tensor $X^{(i+1)}$ with $c_{i+1}$ feature maps.

\subsubsection{Subspace clustering.}

To prune the channels of feature maps from $c_i$ to desired $c_i^{\prime}$ $(0\le c_i^{\prime} \le c_i)$, we cluster the $c_i$ feature maps into $c_i^{\prime}$ clusters. Since different feature maps are generated from the same image, and convolutional filter is a linear operation, the output feature maps of the image will be linearly dependent in different subspaces which satisfy subspace distribution, i.e., one feature map can be expressed in a subspace as a linear combination of other feature maps in the same subspace. This property is called self-expressiveness. We leverage a subspace clustering algorithm~\cite{elhamifar2012Sparse} to cluster the feature maps. Mathematically, this idea is formalized as an optimization problem
\begin{equation}\label{opt}
\underset{C}{\min} \ \|C\|_1 \ s.t. \ X=XC, \ (diag(C)=0),
\end{equation}
where $X\in \mathbb{R}^{HW\times c_i}$ is reshaped $X^{(i)}$, and $C$ is a self-expressiveness matrix. The subspace clustering algorithm is summarized in Algorithm~\ref{algo1}.

\begin{algorithm}[h]
\label{algo1}
 \SetAlgoLined
 \KwIn{The input feature maps $X^{(i)}\in \mathbb{R}^{c_i\times H\times W}$, the desired number of input channels of filter, $c_i^{\prime}$.}
 \textbf{Steps:}\\
 1. Reshape $X^{(i)}$ to $X\in \mathbb{R}^{HW\times c_i}$.\\
 2. Learn the self-expressiveness matrix $C$ from Eq.~(\ref{opt}).\\
 3. Construct an affinity matrix by $W=|C|+|C^T|$.\\
 4. Calculate the Laplacian matrix $L$ of $W$. \\
 5. Calculate the eigenvector matrix $V$ of $L$ corresponding to its $c_i^{\prime}$ smallest nonzero eigenvalues.\\
 6. Perform k-means clustering algorithm on the rows of $V$.\\
 \noindent\KwOut{The clustering result of $X^{(i)}$ with $c_i^{\prime}$ clusters.}
\caption{Subspace Clustering.}
\end{algorithm}

\subsubsection{Filter clustering.}

After clustering $c_i$ feature maps into $c_i^{\prime}$ clusters, we represent the indices of each cluster as $I_1, I_2, \dots, I_{c_i^{\prime}}$. Then, we can prune the channels of filter in layer $i$ by calculating the average channel of each cluster. For the $m$-th filter $W_m^{(i)}$, the average channel can be calculated through the clustering index
\begin{equation}\label{average1}
V_j^{(i)}=\frac{1}{|I_j|}\sum_{p\in I_j} W_{m,p}^{(i)}, \ j=1,2,\dots,c_i^{\prime}
\end{equation}
where $W_{m,p}^{(i)}$ is the $p$-th channel of filter $W_m^{(i)}$, $|I_j|$ is the number of elements in $I_j$. Then the pruned $m$-th filter $W_m^{\prime(i)}$ is the concatenation of $V_j^{(i)}$, $j=1,2,\dots,c_i^{\prime}$. For $m=1,2,\dots,c_{i+1}$, we can obtain their pruned filters using Eq.~(\ref{average1}). Then the pruned filters for layers $i$ are $W^{\prime(i)}=[W_1^{\prime(i)} \ W_2^{\prime(i)} \ \dots \ W_{c_{i+1}}^{\prime(i)}]\in \mathbb{R}^{c_{i+1}\times c_i^{\prime}\times k_h\times k_w}$.

Naturally, the filters of upper layer $i-1$ that produce feature maps $X^{(i)}$ can also be clustered
\begin{equation}\label{average2}
W_j^{\prime(i-1)}=\frac{1}{|I_j|}\sum_{p\in I_j} W_{p}^{(i-1)}, \ j=1,2,\dots,c_i^{\prime}
\end{equation}
where $W_{p}^{(i-1)}$ is the $p$-th filters of $W^{(i-1)}$, $|I_j|$ is the number of elements in $I_j$. The result is $W^{\prime(i-1)}=[W_1^{\prime(i-1)} \ W_2^{\prime(i-1)} \ \dots \ W_{c_i^{\prime}}^{\prime(i-1)}]\in \mathbb{R}^{c_i^{\prime}\times c_{i-1}\times k_h\times k_w}$, where $c_{i-1}$ is the number of filters in layer $i-1$.

\subsubsection{Reconstruction error minimization.}

We reconstruct the output feature maps $X^{(i+1)}$ with pruned filters $W^{\prime(i)}$ by linear least squares. This problem can be formulated as:
\begin{equation}\label{reconstruction}
\min_{W^{\prime(i)}}\|X^{(i+1)}-X^{\prime(i)}*W^{\prime(i)}\|_F^2,
\end{equation}
where $\|\cdot\|_F$ is the Frobenius norm, $*$ is the convolution operation. $X^{\prime(i)}=X^{(i-1)}*W^{\prime(i-1)}\in \mathbb{R}^{c_i^{\prime}\times H\times W}$ are the feature maps produced by the pruned layer $i-1$. The complete filter pruning process is summarized in Algorithm~\ref{algo2}.

\begin{algorithm}
\label{algo2}
 \SetAlgoLined
 \KwIn{The original convolutional filters $W^{(i-1)}$ and $W^{(i)}$, the indices of clustering result $I_1, I_2, \dots, I_{c_i^{\prime}}$.}
 \textbf{Steps:}\\
 1. For layer $i$, calculate the aggregated channel for each filter through the clustering indices, $W^{\prime(i)}\in \mathbb{R}^{c_{i+1}\times c_i^{\prime}\times k_h\times k_w}$.\\
 2. For layer $i-1$, calculate the aggregated filter for each cluster through the clustering indices, $W^{\prime(i-1)}\in \mathbb{R}^{c_i^{\prime}\times c_{i-1}\times k_h\times k_w}$.\\
 3. Minimize the reconstruction error between the original output and the pruned output of layer $i$ by Eq.~(\ref{reconstruction}).\\
 \noindent\KwOut{The pruned convolutional filters $W^{\prime(i-1)}$ and $W^{\prime(i)}$.}
\caption{Filter Pruning.}
\end{algorithm}

\subsection{Pruning Strategy}

The network architectures can be divided into two types, the traditional single path and multi-path convolutional architectures. AlexNet~\cite{krizhevsky2012imagenet} or VGGNet~\cite{simonyan2014very} is the representative for the former, while the latter mainly includes some recent networks equipped with some novel structures like Inception in GoogLeNet~\cite{szegedy2015going} or residual blocks in ResNet~\cite{he2016deep}.

\begin{figure}[t]
\centering
\includegraphics[width=0.9\linewidth]{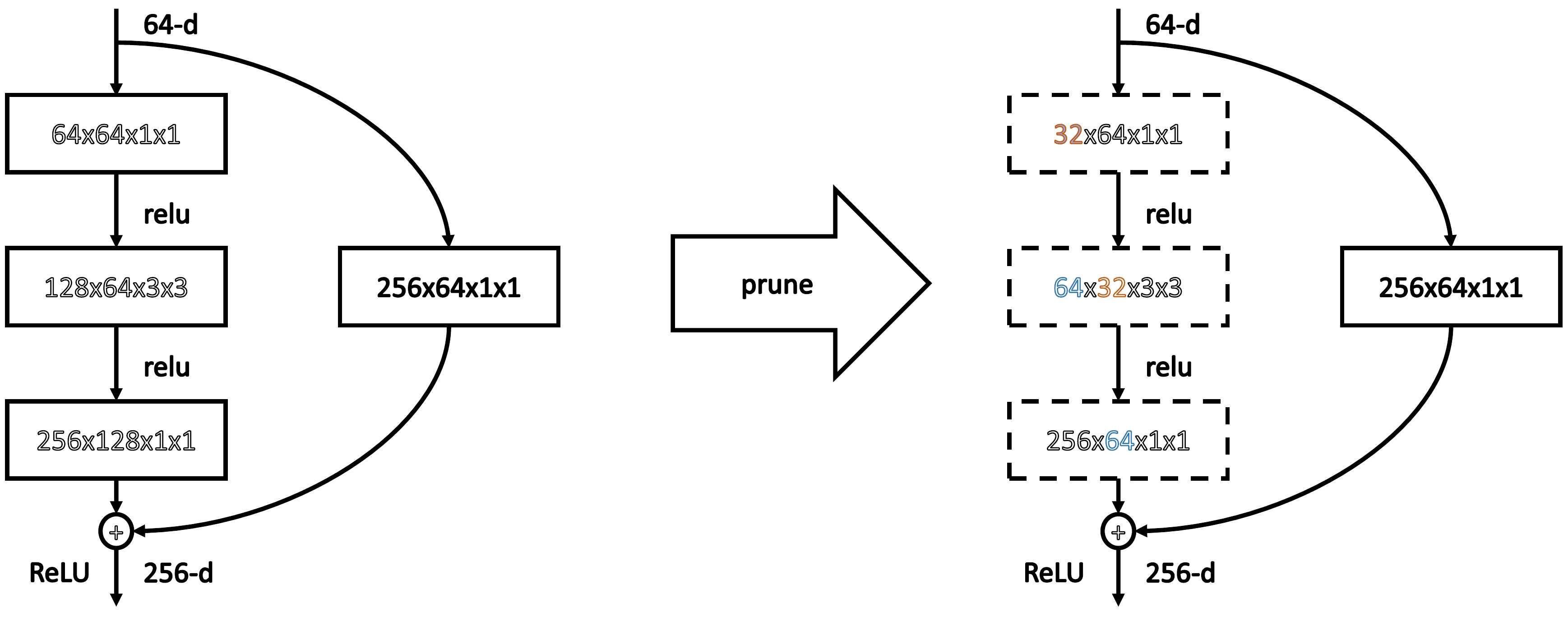}
\caption{Illustration of the ResNet pruning strategy. For each residual block, we only prune the first two convolutional layers, keeping the block output dimension unchanged.}
\label{resnet}
\end{figure}

We use different strategies to prune these two types of networks. For VGG-16, we apply the single layer pruning strategy to the convolutional layer step by step. For ResNet, some restrictions are incurred due to its special structure. For example, the channel numbers of the residual learning branch and the identity mapping branch in the same block need to be consistent in order to finish the sum operation. Thus it is hard to directly prune the last convolutional layer in the residual learning branch. Since most parameters appear in the first two layers, pruning the first two layers is a feasible option which is illustrated in Figure~\ref{resnet}.

\section{Experiment}

Our method is tested on combinations of three popular CNN models with three benchmark datasets: VGG-16~\cite{simonyan2014very} on ILSCVR-12~\cite{russakovsky2015imagenet} and PASCAL VOC 2007~\cite{everingham2008pascal}, ResNet-50~\cite{he2016deep} on ILSCVR-12~\cite{russakovsky2015imagenet}, and CMU-pose~\cite{cao2017realtime} on MSCOCO-14~\cite{Lin2014Microsoft}.

Firstly, we compare several filter selection strategies including ours by pruning single layer for VGG-16~\cite{simonyan2014very} on ILSCVR-12 to exhibit efficiency of our algorithm, followed by whole model pruning for VGG-16~\cite{simonyan2014very}. Secondly, we show the performance of pruning the network with residual architecture, for which ResNet-50~\cite{he2016deep} is selected. Finally, we apply our method to Faster R-CNN~\cite{ren2015faster} and CMU-Pose~\cite{cao2017realtime} to evaluate the generalization capability of our algorithm to challenge visual tasks of object detection and human pose estimation. All the experiments were implemented within Caffe~\cite{jia2014caffe}.

The performance of ConvNets compression is evaluated with different speed-up ratios. Assume that $c$ is the number of filters in the original layer and $c^{\prime}$ is that of the pruned layer, then
\begin{equation}
\text{speed-up ratio} = \frac{c}{c^{\prime}}
\end{equation}

\subsection{Experiments on VGG-16}

VGG-16~\cite{simonyan2014very} is a classic single path CNN with 13 convolutional layers, which has been widely used in vision tasks as a powerful feature extractor. We use single layer pruning and whole model pruning to evaluate the efficiency of our method. The effectiveness is measured by the decrease of top-5 accuracy on validation dataset. The top-5 accuracy of VGG-16~\cite{simonyan2014very} on ILSCVR-12~\cite{russakovsky2015imagenet} validation dataset is $89.9\%$.

\subsubsection{Single layer pruning.}

\begin{figure}[t]
\centering
\begin{minipage}{0.3\linewidth}
  \centerline{\includegraphics[width=4cm]{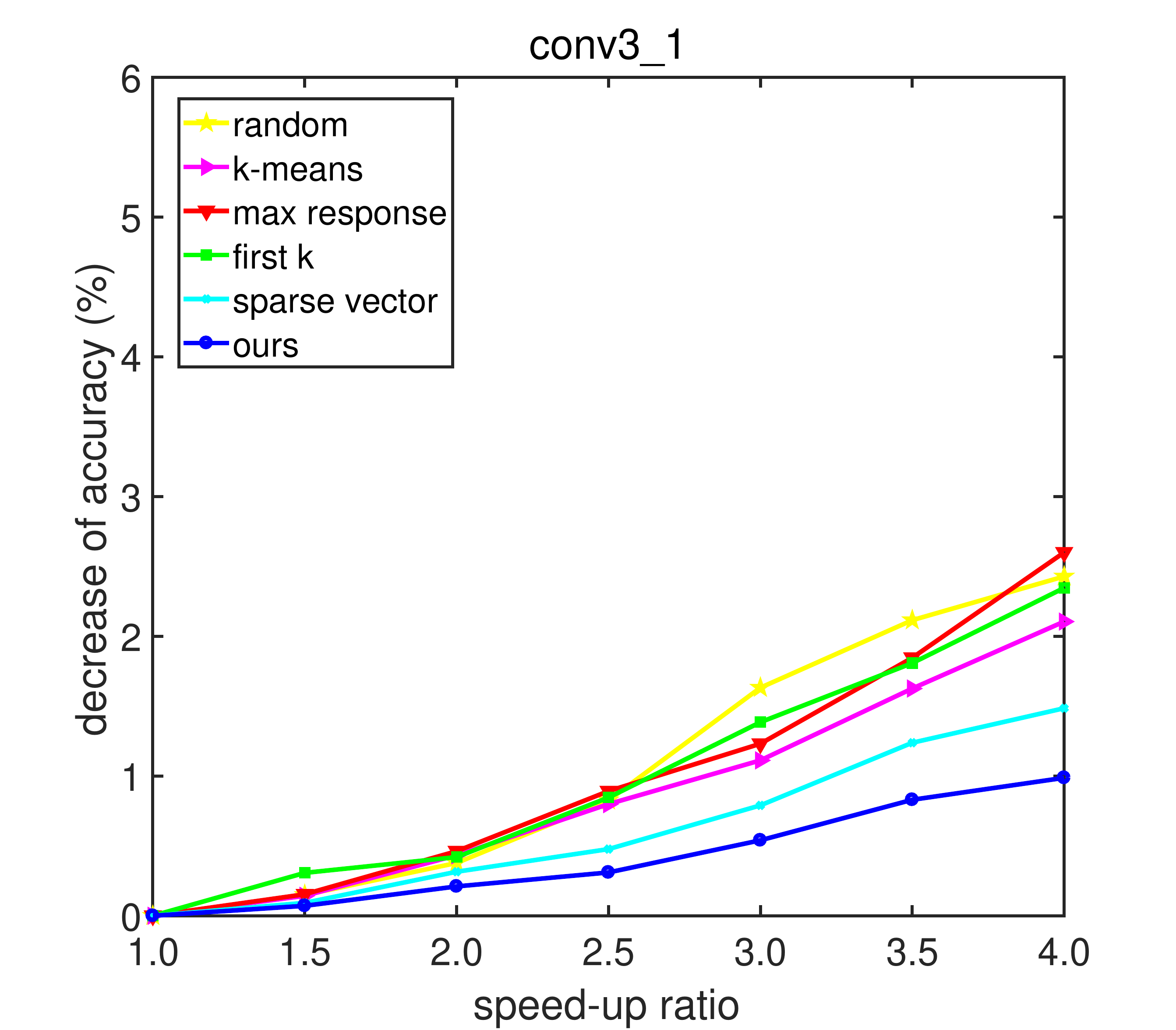}}
\end{minipage}
\begin{minipage}{0.3\linewidth}
  \centerline{\includegraphics[width=4cm]{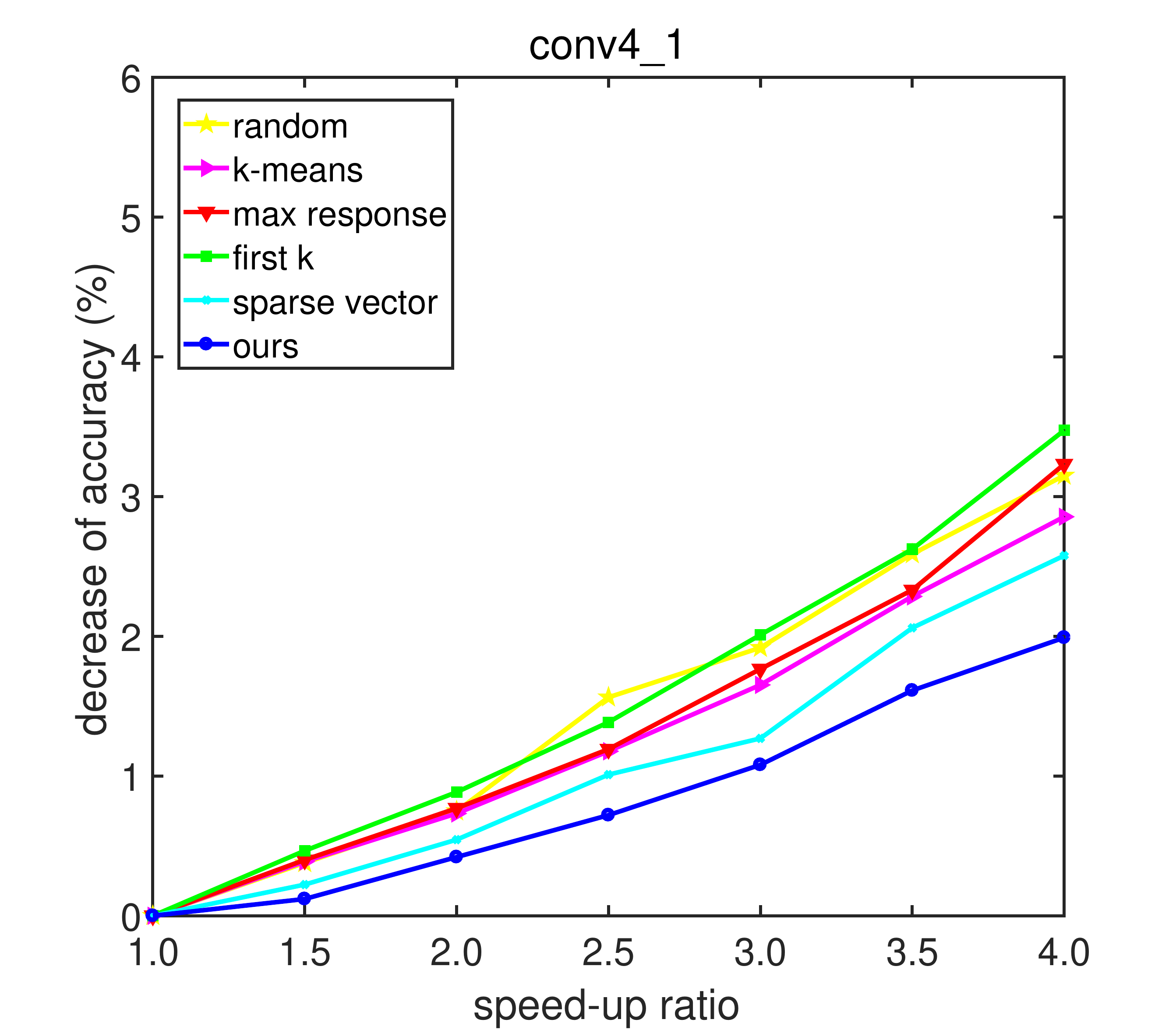}}
\end{minipage}
\begin{minipage}{0.3\linewidth}
  \centerline{\includegraphics[width=4cm]{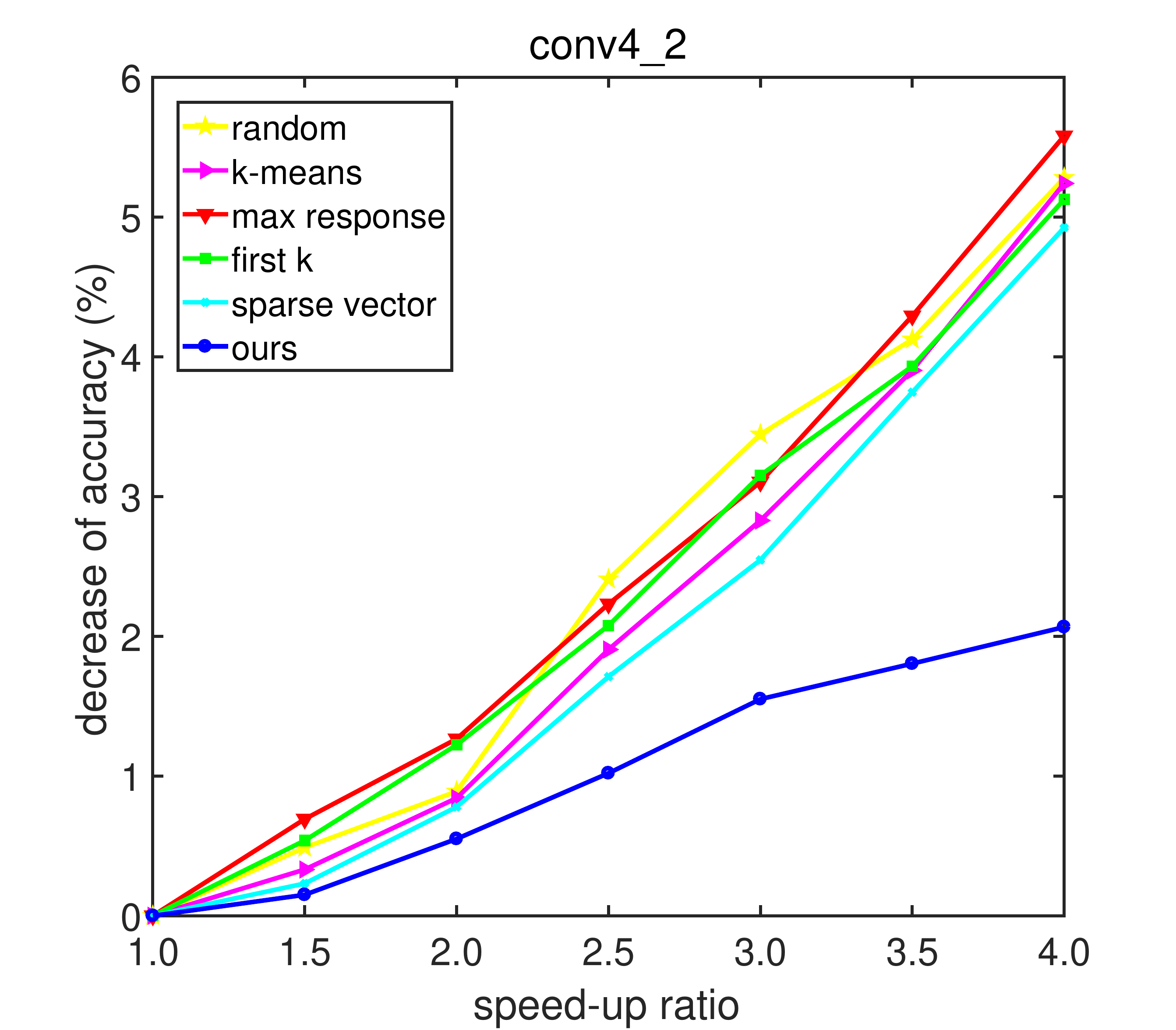}}
\end{minipage}
\caption{Single layer performance analysis under different speed-up ratios (without fine-tuning), measured by decrease of top-5 accuracy on ILSCVR-12 validation dataset.}
\label{siglelayer}
\end{figure}

We first evaluate the single layer acceleration performance of our method. We compare our approach with several existing filter selection strategies. \emph{sparse vector}~\cite{he2017channel} preserves filters according to their importance scores learned by a sparsity regularization method. \emph{max response}~\cite{li2016pruning} selects channels based on corresponding filters that have high absolute sum of weights. To differentiate our approach from the common clustering algorithms, we also select \emph{kmeans} as a baseline. In addition, to validate the necessity of elaborative hand-crafted filter selection algorithm, we also take two naive criteria into consideration. \emph{first k} selects the first k channels. \emph{random} randomly selects a fixed amount of filters. After filter pruning, feature maps reconstruction is implemented without the fine-tuning step. The effectiveness of the methods is measured by reduction of top-5 accuracy on the validation dataset after the reconstruction procedure is accomplished.

Similar to~\cite{he2017channel}, we extracted 10 samples per class, i.e. a total of $10000$ images, to prune channels and minimize reconstruction errors. Images were resized such that the shorter dimension is 256. Then $224 \times 224$ random cropping was applied and the resulting image patches were fed into the network. The testing was made on a crop of $224 \times 224$ pixels at the center of the image. The self-expressiveness matrices for convolutional layers were learned with mini-batch size of 128 and the learning rate varied from $1e^{-3}$ to $1e^{-5}$ in $15$ epochs. After pruning filters, we used a batch size of 64 and varied the learning rate from $1e^{-3}$ to $1e^{-5}$ to minimize the reconstruct error until the loss did not drop continuously. All parameters were optimized with Adam~\cite{kingma2014adam}. We pruned three convolutional layers, i.e., conv3\_1, conv4\_1 and conv4\_2, with aforementioned methods including ours under several speed-up ratios. The results are shown in Figure~\ref{siglelayer}.

As expected, the loss on accuracy is proportional to the speed-up ratio, i.e., error increases as speed-up ratio increases. With the same speed-up ratio, our approach consistently outperforms other methods in different convolutional layers under different speed-up ratios. This shows that our subspace clustering based pruning method can retain more representative information. This enables the feature maps to be reconstructed more effectively. Although the key idea of the \emph{kmeans} option is also clustering, it can not explore the linear relationship between feature maps, obtaining a coarse clustering result. Nevertheless, the performance of \emph{kmeans} is consistently better than the two naive approaches, indicating clustering based pruning strategy is feasible in practice. \emph{max response} performs with high loss of accuracy, sometimes even worse than \emph{first k}. This is probably because \emph{max response} ignores correlations between different filters. The \emph{random} selection option shows good performance, even better than the heuristic methods in some cases. However, this method is not robust in feature maps reconstruction, making it not applicable in practice. In summary, the naive pruning strategies have shown some weakness, which implies that proper filter selection is vital for filter pruning.

It is also noticeable that pruning gradually becomes more difficult from shallow to deep layers. It indicates that whereas shallow layers have much more redundancy, deeper layers make more contribution to the final performance, which is consistent with the observation in~\cite{zhang2016accelerating} and~\cite{he2017channel}. This means it is preferable to prune more parameters in shallow layers rather than deep layers to accelerate the model. Moreover, Figure~\ref{siglelayer} shows that our filter pruning method leads to smaller increase of error compared with other strategies when the deeper layers are compressed.

\subsubsection{Whole model pruning.}

\setlength{\tabcolsep}{16pt}
\begin{table}[t]
  \centering
  \caption{Accelerating the VGG-16 model using a speedup ratio of $2\times$, $4\times$, and $5\times$ respectively. The results show decreases of top-5 validation accuracy (1-view, baseline $89.9\%$).}
  \begin{center}
    \begin{tabular}{c|c|c|c}
    \hline
    Method&$2\times$&$4\times$&$5\times$\cr
    \hline
    Jaderberg \emph{et al.}~\cite{jaderberg2014speeding}& - & 9.7 & 29.7 \cr
    Asym.~\cite{zhang2016accelerating}& 0.28 & 3.84 & - \cr
    Filter pruning~\cite{li2016pruning} (fine-tuned)& 0.8 & 8.6 & 14.6 \cr
    He \emph{et al.}~\cite{he2017channel} (without fine-tune)& 2.7 & 7.9 & 22.0 \cr
    Ours (without fine-tune)& \textbf{2.6} & \textbf{3.7} & \textbf{8.7} \cr
    He \emph{et al.}~\cite{he2017channel} (fine-tuned)& 0 & 1.0 & 1.7 \cr
    Ours (fine-tuned)& \textbf{0} & \textbf{0.5} & \textbf{1.1} \cr
    \hline
    \end{tabular}
    \end{center}
\label{table1}
\end{table}

The whole model acceleration results under $2\times$, $4\times$, $5\times$ are demonstrated in Table~\ref{table1}. Firstly, we applied our approach layer by layer sequentially. Then, our pruned model was fine-tuned for 10 epoches with a fixed learning rate $1e^{-5}$ to gain a higher accuracy. We augmented the data by random cropping of $224 \times 224$ pixels and mirror the cropped patch. Other parameters were the same as in our single layer pruning experiment. Since the last group of convolutional layers (i.e., conv$5\_$x) affects the classification more significantly, these layers were not pruned. After the filter pruning and reconstruction, our approach outperforms the \emph{sparse vector} method~\cite{he2017channel} by a large margin, which is consistent with the results of single layer analysis. In addition, our approach produces more compact models since we do not have the constraint on remaining channels ratios for shallow layers (conv1\_x to conv3\_x) and deep layers (conv4\_x) as required in~\cite{he2017channel}. After fine-tuning, our method achieves $2\times$ speed-up without decrease of accuracy. Under $4\times$ and $5\times$, the accuracy of our method only drops by $0.5\%$ and $1.1\%$ respectively. Our approach outperforms the state-of-the-art filter level pruning approaches (\cite{li2016pruning} and~\cite{he2017channel}). This is because our method retains as much representative information as possible by exploring linear relationship between feature maps via subspace clustering, thus, recovering better approximation to the original data in the subsequent output volume.

\subsection{Experiments on ResNet}

We also tested our method on the recently proposed multi-path network ResNet~\cite{he2016deep}. We selected ResNet-50 as a representation of the ResNet family. During the implementation, we merged batch normalization~\cite{ioffe2015batch} into convolutional weights. This does not affect the outputs of the networks, so that each convolutional layer is followed by ReLU~\cite{nair2010rectified}. Since ResNet-50 consists of residual blocks, we pruned each block step by step, i.e., we pruned ResNet-50 from block 2a to 5c sequentially. In this experiment, for each block, we only pruned the convolutional layers that learned the residual mapping. Therefore, we only pruned the first two layers of each block in ResNet-50 for simplicity, leaving the block output and projection shortcuts unchanged. Pruning these parts may lead to further compression, but can be quite difficult if not entirely impossible. We leave this exploration as a future work. After each block had been pruned, we used Adam~\cite{kingma2014adam} with mini-batch size of 64 and varied the learning rate from $1e^{-3}$ to $1e^{-5}$ to minimize reconstruction error until the loss did not drop continuously. The model was fine-tuned in 20 epochs with fixed learning rate $1e^{-5}$ to gain a higher accuracy.

\setlength{\tabcolsep}{16pt}
\begin{table}[t]
  \centering
  \caption{$2\times$ acceleration for ResNet-50 on ILSCVR-12. The results show decrease from the baseline networks top-5 accuracy of $92.2\%$ (one view).}
  \begin{center}
    \begin{tabular}{c|c}
    \hline
    Method&Increased err.\cr
    \hline
    He \emph{et al.}~\cite{he2017channel}& 8.0 \cr
    Ours(without fine-tune)& \textbf{5.2} \cr
    He \emph{et al.}~\cite{he2017channel}(enhanced)& 4.0 \cr
    He \emph{et al.}~\cite{he2017channel}(enhanced, fine-tuned)& 1.4 \cr
    Ours(fine-tune)& \textbf{1.0} \cr
    \hline
    \end{tabular}
    \end{center}
\label{table2}
\end{table}

The results of $2\times$ acceleration on ResNet-50 are presented in Table~\ref{table2}. Our approach outperforms the state-of-the-art method~\cite{he2017channel} both before or after the fine-tuning. In addition, while pruning ResNet-50, He \emph{et al.}~\cite{he2017channel} kept 70\% and 30\% channels for sensitive residual blocks and other blocks respectively. Our approach, without these constraints, is simpler and more efficient. Our pruning strategy can obtain more representative filters by eliminating redundancy in feature map subspace, enabling the reconstruct error to be better minimized.

\subsection{Generalization Capability of the Pruned Model}

To explore the generalization capability of our method, we ran experiments on two challenging vision tasks: object detection and human pose estimation. We used Faster R-CNN~\cite{ren2015faster} on PASCAL VOC 2007 for the former task and CMU-pose~\cite{cao2017realtime} on MSCOCO14~\cite{Lin2014Microsoft} for the latter one. Both networks were accelerated by our approach under $2\times$ and $4\times$ speed-up ratios. The performance is evaluated in terms of mean Average Precision (mAP).

\subsubsection{Acceleration for object detection.}

\setlength{\tabcolsep}{16pt}
\begin{table}[t]
  \centering
  \caption{$2\times$ and $4\times$ acceleration on Faster R-CNN detection.}
  \begin{center}
    \begin{tabular}{c|c|c}
    \hline
    Speedup&mAP&$\Delta$mAP\cr
    \hline
    Baseline& 68.7 & - \cr
    He \emph{et al.}~\cite{he2017channel} ($2\times$) & 68.3 & 0.4 \cr
    Ours ($2\times$) & \textbf{68.5} & \textbf{0.2} \cr
    He \emph{et al.}~\cite{he2017channel} ($4\times$) & 66.9 & 1.8 \cr
    Ours ($4\times$) & \textbf{67.7} & \textbf{1.0} \cr
    \hline
    \end{tabular}
    \end{center}
\label{table3}
\end{table}

For convenience, we compressed the Faster R-CNN model with VGG-16 as its backbone. Since there is no redundancy in the convolutional layers except those in VGG-16, we used the same parameters as in our VGG-16 experiment. To compare with the alternative approach fairly, we followed the setting as in~\cite{he2017channel}. We first performed channel pruning on VGG-16 on the ImageNet. Then we used the pruned model as the pre-trained model for Faster R-CNN. The model acceleration is demonstrated on the PASCAL VOC 2007 object detection benchmark~\cite{everingham2008pascal} which contains 5k training images and 5k testing images. From Table~\ref{table3}, we observe $0.2\%$ mAP drop with our $2\times$ model, which outperforms the method of He \emph{et al.}~\cite{he2017channel}. Such small mAP drop will not generate significant negative effect in real applications, but brings much benefit in efficiency and model complexity reduction.

\subsubsection{Acceleration for human pose estimation.}

\setlength{\tabcolsep}{16pt}
\begin{table}[t]
  \centering
  \caption{$2\times$ and $4\times$ acceleration on CMU-Pose human pose estimation.}
  \begin{center}
    \begin{tabular}{c|c|c}
    \hline
    Speedup&mAP&$\Delta$mAP\cr
    \hline
    Baseline& 57.6 & - \cr
    $2\times$& 56.8 & 0.8 \cr
    $4\times$& 55.7 & 1.9 \cr
    \hline
    \end{tabular}
    \end{center}
\label{table4}
\end{table}

CMU-pose~\cite{cao2017realtime} is a bottom-up approach for multi-person 2D pose estimation. It simultaneously predicts heat maps and part affinity fields (PAFs) for body parts and body limps respectively. Then it joins two corresponding detection results into the same group by using an associated PAF. The architecture of the network consists of two parts. The first 10 layers of VGG-19~\cite{simonyan2014very} is used as the first part of the network to extract features. In the second part, the network is split into two branches: one branch predicts the confidence maps, and the other predicts the affinity fields. Each branch is an iterative prediction architecture, which refines the predictions over successive stages with intermediate supervision at each stage. Since there is no redundancy in the last convolutional layer of each stage (i.e., conv5\_5\_CPM\_Lx and Mconv7\_stagex\_Lx), we pruned the rest convolutional layers in the same manner as in the single layer pruning strategy. Similar to our VGG-16 experiment, we randomly selected $10000$ samples for filter pruning and reconstruction. After pruning and reconstruction, the model was fine-tuned in $15$ epoches with a fixed learning rate $1e^{-5}$. Other parameters were the same as in our VGG-16 pruning experiment. CMU-pose model compression results under $2\times$ and $4\times$ are demonstrated in Table~\ref{table4}. The results show 0.8\% mAP drop of our $2\times$ model, which showcase the effectiveness of our method.

\section{Conclusion}

Current deep CNNs are accurate with high inference costs. In this paper, we have presented a novel filter pruning method for deep neural networks. Since it is observable that there is linear relationship in different feature map subspaces, we can eliminate the redundancy in convolutional filters by applying subspace clustering on feature maps. Different from existing filter pruning methods which directly remove filters based on their importance, our approach better retrieves the representative information according to the linear relationship in feature map subspaces, so most important information can be retained by the mean of each cluster. Our method only requires off-the-shelf libraries. The reduced CNNs are inference efficient networks while maintaining accuracy. Compelling speed-up and accuracy are demonstrated on both VGG-Net and ResNet with ILSCVR-12. Moreover, experiments on other computer vision tasks also show the feasibility of our compression method in practice.

\bibliographystyle{splncs03}
\bibliography{egbib}
\end{document}